\documentclass[letterpaper]{article} 
\usepackage{aaai25}  
\usepackage{times}  
\usepackage{helvet}  
\usepackage{courier}  
\usepackage[hyphens]{url}  
\usepackage{graphicx} 
\usepackage{natbib}  
\usepackage{caption} 
\usepackage{algorithm}
\usepackage{algorithmic}

\usepackage{amsfonts}
\usepackage{amsmath}
\usepackage{booktabs}
\usepackage{multirow} 
\usepackage{tabularx}
\usepackage{adjustbox}

\usepackage{newfloat}
\usepackage{listings}
\DeclareCaptionStyle{ruled}{labelfont=normalfont,labelsep=colon,strut=off} 
\floatstyle{ruled}
\newfloat{listing}{tb}{lst}{}
\floatname{listing}{Listing}
\title{PSReg: Prior-guided Sparse Mixture of Experts for Point Cloud Registration}
\author{
    Xiaoshui Huang\textsuperscript{\rm 1,5}\equalcontrib, 
    Zhou Huang\textsuperscript{\rm 2,6}\equalcontrib, 
    Yifan Zuo\textsuperscript{\rm 2,6}\thanks{Corresponding author.}, 
    Yongshun Gong\textsuperscript{\rm 3},\\
    Chengdong Zhang\textsuperscript{\rm 1},
    Deyang Liu\textsuperscript{\rm 4}, 
    Yuming Fang\textsuperscript{\rm 2,6}
}
\affiliations{
    \textsuperscript{\rm 1}Shanghai Jiao Tong University,
    
    \textsuperscript{\rm 2}Jiangxi University of Finance and Economics, 
 
    \textsuperscript{\rm 3}Shandong University, 
    
    \textsuperscript{\rm 4}Anqing Normal University,
    
    \textsuperscript{\rm 5}Huaihua University, 
    
    \textsuperscript{\rm 6}Jiangxi Provincial Key Laboratory of Multimedia Intelligent Processing
    
}

\begin{document}

\maketitle
\begin{abstract}
The discriminative feature is crucial for point cloud registration. Recent methods improve the feature discriminative by distinguishing between non-overlapping and overlapping region points. However, they still face challenges in distinguishing the ambiguous structures in the overlapping regions. Therefore, the ambiguous features they extracted resulted in a significant number of outlier matches from overlapping regions. To solve this problem, we propose a prior-guided SMoE-based registration method to improve the feature distinctiveness by dispatching the potential correspondences to the same experts. Specifically, we propose a prior-guided SMoE module by fusing prior overlap and potential correspondence embeddings for routing, assigning tokens to the most suitable experts for processing. In addition, we propose a registration framework by a specific combination of Transformer layer and prior-guided SMoE module. The proposed method not only pays attention to the importance of locating the overlapping areas of point clouds, but also commits to finding more accurate correspondences in overlapping areas. Our extensive experiments demonstrate the effectiveness of our method, achieving state-of-the-art registration recall (95.7\%/79.3\%) on the 3DMatch/3DLoMatch benchmark. Moreover, we also test the performance on ModelNet40 and demonstrate excellent performance.
\end{abstract}

\section{Introduction}
Point cloud registration is fundamental in computer vision and widely applies in domains such as 3D reconstruction \cite{gross2019alignnet, izadi2011kinectfusion}, autonomous driving \cite{nagy2018real}, and augmented reality \cite{huang2016real}. The objective is to estimate the optimal rigid transformation between two point clouds, thereby establishing their correspondences within the same coordinate system. Currently, learning-based methods are leading the task of point cloud registration. However, successfully registering them remains challenging when the point clouds only partially overlap.

\begin{figure}[t]
	\centering
	\includegraphics[width=\linewidth]{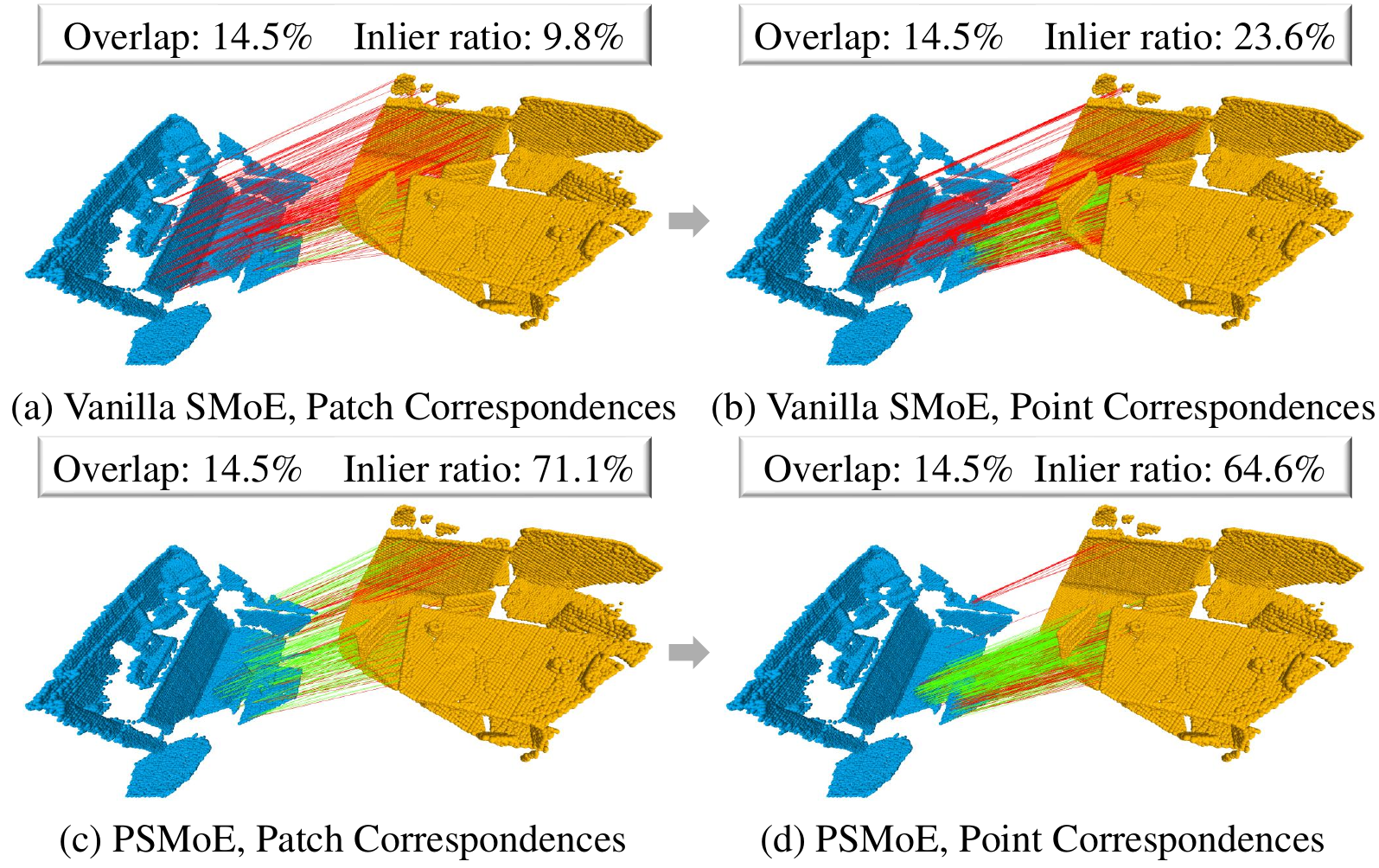}
	\caption{Performance gaps in applying vanilla SMoE and our PSMoE. Fig. (a) and (c) show coarse patch correspondences, (b) and (d) show fine point correspondences.}
	\label{fig:correspondences}
\end{figure}

Learning-based works on partial overlap point cloud registration can be classified into two categories: non-Transformer-based methods and transformer-based methods. Non-Transformer-based methods \cite{zeng20173dmatch, huang2020feature, choy2019fully, ao2021spinnet} directly utilize existing networks for single-stage feature extraction and estimate the transformation matrix. However, they face challenges in integrating global information and feature alignment. In comparison, the Transformer-based methods \cite{wang2019deep, huang2021predator,huang2022imfnet, yu2021cofinet,mei2022overlap, yew2022regtr, qin2022geometric} utilize a Transformer module to further enhance the features extracted by the existing backbones. Nevertheless, in scenarios with a large number of similar structures or textureless structures, these methods face the issue of ambiguous matching. Among them, PEAL \cite{yu2023peal} explicitly encodes intra-frame overlapping and non-overlapping regions using prior information, which alleviates the above ambiguous matching to a certain extent. However, if the overlapping region between the two point clouds contains a large amount of ambiguous structures (floor or wall without obvious texture), PEAL still faces difficulty. This is because it cannot distinguish the ambiguous structures within the overlapping regions, resulting in a large number of outlier matches.

To address the aforementioned issues, we believe that applying sparse mixture of experts (SMoE) in Transformer is a feasible solution. The routing network of SMoE can cluster input tokens into different groups and then distribute these groups to different experts for processing. If the routing network can distinguish between overlapping and non-overlapping regions of tokens, and further capture the correspondence between tokens of two point clouds, then the multi-expert networks through SMoE can extract more discriminative features. To the best of our knowledge, we are pioneering the exploration of SMoE for addressing the point cloud registration task.

However, SMoE's simple application does not fulfill our expectations. Due to insufficient interaction between the features of the source and target point clouds, and more importantly, the lack of relevant guidance, it's challenging for SMoE's routing network to differentiate tokens from overlapping and non-overlapping regions (Fig. \ref{fig:correspondences} (a)). Inspired by PEAL \cite{yu2023peal}, we propose \textbf{P}rior-guided routing \textbf{S}parse \textbf{M}ixture \textbf{o}f \textbf{E}xperts (PSMoE), a novel module to address the aforementioned challenge. Specifically, we first calculate superpoint correspondences using transformations estimated by a SOTA pretrained model. Then, we design a Prior superpoint Correspondences Encoding (PCE) module to obtain prior embeddings. PCE module encodes matched superpoints with discrete ordered numbers, while non-matched superpoints are assigned the same number, all the numbers assigned to superpoints are encoded using an embedding layer. Finally, the routing network groups the original tokens based on prior embeddings and sends them to selected experts to extract distinctive features. Unlike PEAL, our PCE module not only encodes prior overlapping information but also further captures the potential correspondence of matching points. These prior correspondences guide the distribution of potential matching tokens to be processed by the same expert as much as possible, which facilitates feature alignment. Therefore, our method effectively locates overlapping areas and enhances matching accuracy within these regions (Fig. \ref{fig:correspondences} (c)).

We design a coarse-to-fine point cloud registration framework based on the PSMoE module, named \textbf{P}rior-guided \textbf{S}MoE-based \textbf{Reg}istration (PSReg), to estimate the final transformation matrix. Additionally, we conduct extensive experiments on indoor benchmarks and synthetic datasets, demonstrating the superiority of our approach over previous methods. A comprehensive series of ablation studies validate the effectiveness of each module within our proposed methodology. Our contributions can be summarized as follows:

\begin{itemize}

  \item We analyze the reasons for the poor performance of directly applying vanilla SMoE in point cloud registration, and propose a prior-guided SMoE (PSMoE) module to further improve the feature distinctiveness within the overlapping regions. To the best of our knowledge, we are the pioneers in exploring the SMoE in the point cloud registration task.
 
  \item We propose a registration method, PSReg, by combining both the Transformer and our PSMoE module.  
  
  \item We conducted extensive experiments to verify the effectiveness of our methodology and explored the feasibility of applying SMoE to the point cloud registration task.
\end{itemize}

\section{Related Work}

\subsection{Non-Transformer-based Registration}
The overall idea of non-Transformer-based point cloud registration methods is typically using networks specifically designed for 3D data to extract features in a single stage, without further enhancing high-level features. These features are then utilized to establish correspondences and indirectly solve for the transformation matrix, or the transformation matrix is solved directly from the features. Based on the different types of input data, we can further broadly classify the existing methods into two categories. Firstly, methods based on structured data \cite{zeng20173dmatch, elbaz20173d, choy2019fully, ao2021spinnet}, which typically involve projecting the point cloud into 2D images or constructing 3D volume, and directly applying 2D CNN or 3D CNN. Secondly, methods based on raw point clouds \cite{aoki2019pointnetlk,du2019kpsnet, huang2020feature, bai2020d3feat, xu2021omnet, huang2022robust}, which directly use the point cloud data as input. PointNet \cite{qi2017pointnet} directly uses MLPs to extract point-wise features and addresses permutation invariance through max pooling. PointNet++ \cite{qi2017pointnet++} further introduces sampling and grouping layers to capture multi-scale contextual information. KPConv \cite{thomas2019kpconv} uses kernel points that carry convolution weights to simulate the kernel pixels in 2D convolution, thereby defining the convolution operation on raw point clouds. The above three backbone networks, which directly process raw point cloud data, are widely used in point cloud registration. However, the above methods face issues with information loss and challenges in capturing global contextual information.

\subsection{Transformer-based Registration}
Transformer-based registration methods typically involve a two-stage feature extraction process. Initially, high-level features are extracted by the existing 3D backbone, followed by further feature enhancement through Transformer. DCP \cite{wang2019deep} is the pioneer in utilizing Transformer to enhance feature extraction in point cloud registration, whereby unaligned point clouds are fed into a feature embedding module, followed by context aggregation executed by a Transformer encoder. REGTR \cite{yew2022regtr} replaces explicit feature matching and outlier filtering, typically performed by RANSAC, with an end-to-end Transformer framework to directly discover point cloud correspondences. Taking into consideration that the vanilla Transformer does not explicitly encode the geometric structure of point clouds, GeoTransformer \cite{qin2022geometric} is engineered with a geometric self-attention module that expressly captures the internal geometric structures of point clouds. PEAL \cite{yu2023peal} introduces an overlapping prior, utilizing one-way attention to relieving feature ambiguity. However, these methods are prone to failure when dealing with scenes containing a large number of ambiguous structures.

\subsection{Sparse Mixture of Experts (SMoE)}
SMoE is an efficient deep learning model architecture that utilizes a multi-expert system to handle complex tasks. In this structure, the network is composed of multiple experts (multiple FFNs) and a gating mechanism. The gating mechanism is responsible for deciding which experts to activate based on the input data, allowing the network to dynamically adjust its behavior for specific tasks, thereby enhancing processing speed and efficiency. The efficacy of SMoE has been extensively demonstrated across various tasks in NLP \cite{shazeer2017outrageously, lepikhin2020gshard, zhou2022mixture, fedus2022switch, jiang2024mixtral} and CV \cite{gross2017hard, abbas2020biased, pavlitskaya2020using, riquelme2021scaling}. 

However, research on SMoE in 3D vision remains unexplored. In fact, the SMoE is naturally suited for the task of partially overlapping point cloud registration. Partial overlapping point cloud registration focuses more on the overlapping regions of the point clouds. The gating mechanism of SMoE can sparsely activate a subset of experts to specifically handle data from these overlapping regions, which can enhance the uniqueness of the features. The aforementioned sparse activation method is theoretically feasible, but it is difficult to implement in practical applications. Therefore, to achieve controllable sparse activation, we propose a novel prior-guided routing mechanism that leverages SMoE for more effective discriminative feature extraction.

\section{Method}

\subsection{Preliminary and Analysis}
In this section, we will introduce the preliminaries of point cloud registration and sparse mixture of experts (SMoE).
\subsubsection{Point Cloud Registration.} Given a pair of partially overlapping point clouds, namely the source point cloud $\mathcal{P} = \{ p_i \in \mathbb{R}^{3} | i = 1,...,N\}$ and the target point cloud $\mathcal{Q} = \{ q_j \in \mathbb{R}^{3} | j = 1,...,M\}$, where $N$ and $M$ are the number of points in point clouds $\mathcal{P}$ and $\mathcal{Q}$. The goal of point cloud registration is to predict the rotation matrix $R \in SO(3)$ and translation vector $t \in \mathbb{R}^{3}$ to align $\mathcal{P}$ and $\mathcal{Q}$ under the same coordinate system. Our approach needs to be combined with Transformer, while Transformer-based registration methods typically employ an encoder (existing 3D feature extraction networks) for downsampling the raw point clouds and extracting relevant neighborhood features before applying the Transformer. We use $\mathcal{\hat{P}} = \{ \hat{p}_{i} \in \mathbb{R}^{3} | i = 1,...,N^{\prime}\}$ and $\mathcal{\hat{Q}} = \{ \hat{q}_{j} \in \mathbb{R}^{3} | j = 1,...,M^{\prime}\}$ to represent downsampled points (superpoints), and the corresponding superpoint features are represented by $\mathcal{\hat{F}^{P}} \in  \mathbb{R}^{N^{\prime}\times d}$ and $\mathcal{\hat{F}^{Q}} \in  \mathbb{R}^{M^{\prime}\times d}$.

\subsubsection{Vanilla SMoE.}  The vanilla SMoE \cite{riquelme2021scaling} is a neural network layer comprising self-attention mechanisms and multiple feed-forward expert networks. Given an input point cloud token $x$, it is sent to the routing network $\mathcal{G}(\cdot)$, which delegates the token to several highly relevant experts for processing. Typically, the routing network is a fully connected layer, represented as follows:
\begin{equation}
    \mathcal{G} = \textit{top-k}(\textit{softmax}(W_gx)),
\end{equation}
Where $W_g$ represents the trainable parameters in the router neural network and $\textit{top-k}(\cdot)$ select the $k$ experts with the largest gating values. In our approach, $k=1$. The final output of the SMoE layer will be the summary of the weighted active expert features:
\begin{equation}
    y = \mathcal{G}_i \cdot E_i(x),
    \label{eq:smoe_output}
\end{equation}
where $E_i$ represents the neural network of expert i, and $\mathcal{G}_i$ represents the corresponding gate value.

\subsubsection{Analyze Vanilla SMoE.} 
\begin{figure}[t]
	\centering
	\includegraphics[width=\linewidth]{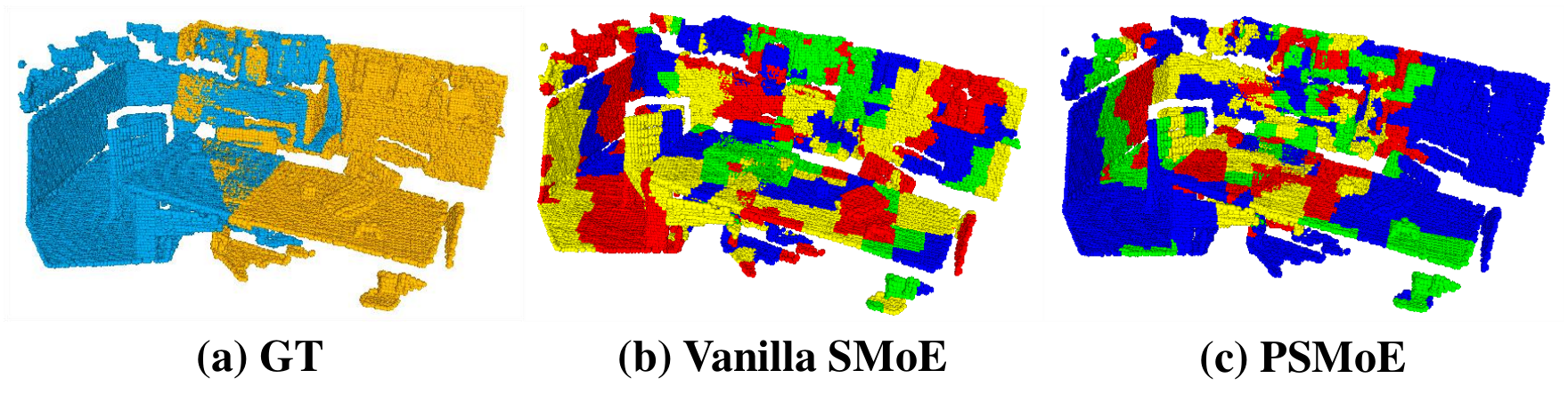}
	\caption{ The visualization of the routing of tokens (patches) in the last SMoE block, where the color of the point cloud patches indicates the selected experts. 
 } 
	\label{fig:token_routing}
\end{figure}

The multi-expert network of SMoE can focus on different features or patterns, which should help enhance the uniqueness of superpoint features in different regions. To analyze the ability of SMoE in the point cloud registration, we directly use vanilla SMoE on the selected baseline model \cite{yu2023peal}. Specifically, we replace the FFN layer in the self-attention block of the baseline with the SMoE layer. This SMoE with the tokens choice routing policy, following the default configurations in \cite{fedus2022switch}. To intuitively observe the routing results of vanilla SMoE, we visualized the distribution of tokens in the last SMoE block. As illustrated in Fig. \ref{fig:token_routing} (b), the tokens do not exhibit a clear clustering phenomenon, indicating that the vanilla SMoE does not handle the expert subsets for tokens in overlapping and non-overlapping regions separately.

\emph{Analysis:} The primary reason for the failure of vanilla SMoE in locating overlapping regions is that the routing network is typically just a fully connected layer. Without explicit guiding signals, its expressive capability struggles to capture the overlapping information between point clouds.

\begin{figure}[h]
	\centering
	\includegraphics[width=\linewidth]{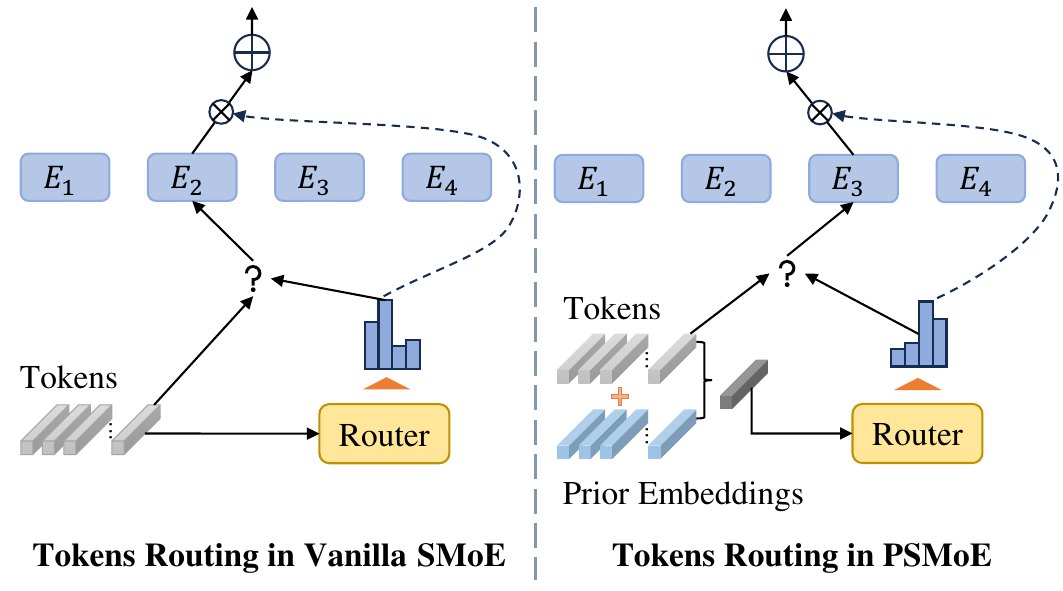}
	\caption{Difference between vanilla SMoE and PSMoE.}
	\label{fig:PSMoE}
\end{figure}

\subsection{PSMoE Module}

To solve the above-mentioned problem and further alleviate matching ambiguities in overlapping areas, we propose \textbf{P}rior-guided \textbf{S}parse \textbf{MoE} (PSMoE), as shown in Fig. \ref{fig:PSMoE}, which offers explicit prior guidance. Fig. \ref{fig:token_routing} (c) shows the token distribution in our method. Due to the introduction of prior information, the routing network can distinguish between overlapping and non-overlapping patches. Therefore, providing clear guidance signals can assist the routing network in better identifying which experts are more suitable for specific inputs. Next, we will sequentially introduce our approach from three aspects: prior superpoint correspondence prediction, prior superpoint correspondence encoding, and prior-guided routing.

\subsubsection{Prior Superpoint Correspondence Prediction.} 
Superpoint correspondences not only provide overlapping information but also directly associate matching superpoints in two point clouds. These explicit signals can effectively guide the routing network as priors to activate proper expert extract discriminative point-wise features. However, it is a chicken-and-egg problem between superpoint correspondences and discriminative point-wise feature extraction.
To solve this problem, inspired by the approach of \cite{yu2023peal}, we calculate the \emph{prior superpoint correspondences} by using a pretrained SOTA registration model. Specifically, we employ GeoTransformer to predict the prior transformation matrix $T_{p} = \{R_{p}, t_{p} \}$, and use $T_{p}$ to compute the overlap ratio matrix $\mathcal{\hat{O}} \in \mathbb{R}^{N^{\prime} \times M^{\prime}}$ between the superpoints (patches) of $\hat{P}$ and $\hat{Q}$. Then, we select entries in $\mathcal{\hat{O}}$ with an overlap ratio greater than $\tau_o$ to serve as prior superpoint correspondences $\mathcal{\bar{C}} = ( \mathcal{P}^{\prime}, \mathcal{Q}^{\prime})$, and the corresponding prior overlap ratios are represented as $\bar{\mathcal{O}}$.

\subsubsection{Prior Superpoint Correspondence Encoding (PCE).} 
To aid the routing network in discerning the precise locations of tokens within overlapping regions and to effectively capture the token correspondences in these putative overlap areas, our aim is to partition all tokens into $L$ (the number of experts) clusters. These clusters can be broadly classified into two categories: the non-anchor cluster (a singular cluster where the majority of tokens reside in non-overlapping zones) and anchor clusters (comprising $L-1$ clusters primarily housing tokens within overlapping regions). Ideally, each sub-cluster within the anchor clusters should encompass tokens with potential matches between the two sets of data points. 

\begin{figure}[]
	\centering
	\includegraphics[width=\linewidth]{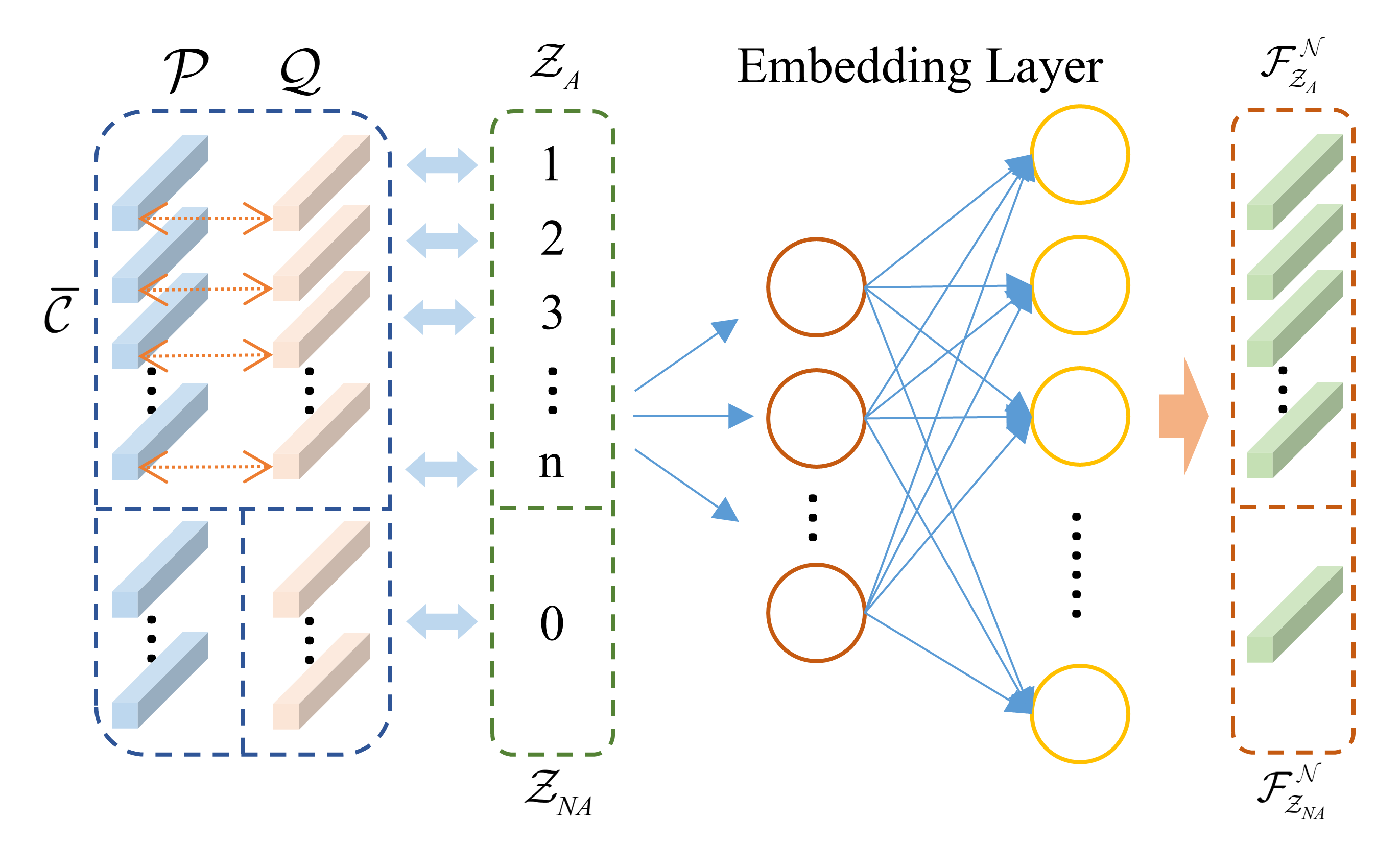}
	\caption{Diagram of PCE module.}
	\label{fig:PCE}
\end{figure}

Consequently, we have devised a PCE module to explicitly encode prior superpoint correspondences, as illustrated in Fig. \ref{fig:PCE}. Specifically, we first encode the prior superpoint correspondences $\mathcal{\bar{C}}$ using a discrete and ordered sequence $\mathcal{Z}_A$, with $\bar{c} = |\mathcal{Z}_A| = |\mathcal{\bar{C}}|$. We then employ the same numerical value $\mathcal{Z}_{NA}$ to encode unmatched superpoints in $\mathcal{P}$ and $\mathcal{Q}$. Finally, we obtain a set of natural numbers, denoted as $\mathcal{N} = \{ i \in \mathbb{N} \mid i = 0,...,\bar{c} \}$:
\begin{equation}
    \mathcal{N} = Concat[\mathcal{Z}_{NA}, \mathcal{Z}_{A}],
\end{equation}
Considering the continuous nature of the sinusoidal embedding function \cite{vaswani2017attention}, we utilize this function and MLP to form an embedding layer to compute the prior embeddings $\mathcal{F}^{\mathcal{N}} \in \mathbb{R}^{(\bar{c}+1) \times d}$ for $\mathcal{N}$:
\begin{equation}
    \begin{cases}
        \begin{aligned}
            \Gamma_{i,2k}^{\mathcal{N}} = \sin(\frac{i}{10000^{2k / d}}) \\
            \Gamma_{i,2k+1}^{\mathcal{N}} = \cos(\frac{i}{10000^{2k / d}}) 
        \end{aligned},
    \end{cases}
    \mathcal{F}^{\mathcal{N}} = MLP(\Gamma^{\mathcal{N}}),
\end{equation}
Based on $\bar{\mathcal{C}}$, we select the corresponding superpoints from $\mathcal{P}$ to form a set of anchor superpoints $\mathcal{A} \subset \hat{\mathcal{P}}$. However, superpoints in $\mathcal{A}$ may correspond to multiple superpoints in $\mathcal{Q}$ that meet the criteria (overlap ratio greater than $\tau_o$), which are denoted as $\mathcal{A}^{\prime}$. Therefore, to ensure the uniqueness of the prior embedding corresponding to $\mathcal{A}^{\prime}$, we perform a weighted sum based on the overlap ratio. For each superpoint in $\mathcal{A}$, we search for the entries corresponding to this superpoint from $\mathcal{P}^{\prime}$ in $\mathcal{\bar{C}}$ and the corresponding index is represented as $\mathcal{X}_i$. Ultimately, the prior embeddings $\ddot{\mathcal{F}}^{\mathcal{P}} \in \mathbb{R}^{|\hat{\mathcal{P}}| \times d}$ corresponding to the superpoints in $\hat{\mathcal{P}}$ is denoted as:

\begin{equation}
    \ddot{\mathcal{F}}^{\mathcal{P}}_i =
    \begin{cases}
        \textit{softmax}(\mathcal{\bar{O}}(\mathcal{X}_i)) \cdot \mathcal{F}^{\mathcal{N}}_{\mathcal{Z}_A}(\mathcal{X}_i), & |\mathcal{X}_i| > 1, \\
        \mathcal{F}^{\mathcal{N}}_{\mathcal{Z}_A}(\mathcal{X}_i), & |\mathcal{X}_i| = 1, \\
        \mathcal{F}^{\mathcal{N}}_{\mathcal{Z}_{NA}}, & otherwise.
    \end{cases}
\end{equation}
Where $\mathcal{F}^{\mathcal{N}}_{\mathcal{Z}_A}$ and $\mathcal{F}^{\mathcal{N}}_{\mathcal{Z}_{NA}}$ respectively represent the embeddings corresponding to $\mathcal{Z}_A$ and $\mathcal{Z}_{NA}$. The prior embeddings $\ddot{\mathcal{F}}^{\mathcal{Q}} \in \mathbb{R}^{|\hat{\mathcal{Q}}| \times d}$ corresponding to $\hat{\mathcal{Q}}$ are obtained in the same manner.
 
\begin{figure*}[t]
	\centering
	\includegraphics[width=\linewidth]{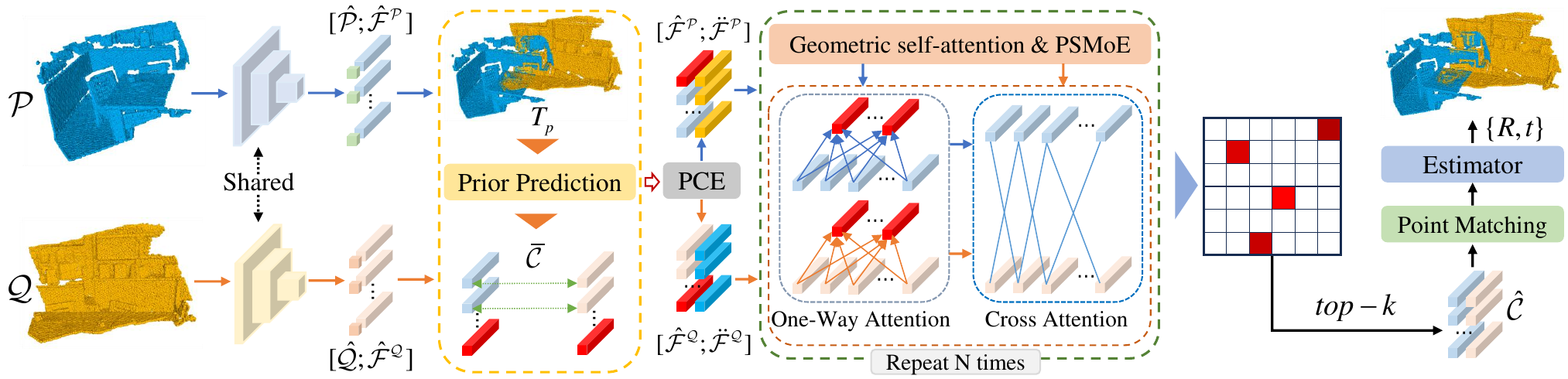}
        \caption{ The framework of PSReg. The red blocks represent superpoints located in the putative non-overlapping areas.}
	\label{fig:frame}
\end{figure*}

\subsubsection{Prior-guided Routing.} 

In Fig. \ref{fig:PSMoE}, it is illustrated that to ensure prior embeddings effectively guide the routing process and simplify model design, we adopt a straightforward yet efficient approach. This involves directly integrating them with prior information before distributing the input tokens through the routing network. More precisely, We sum the prior embeddings encoded by the PCE module with the original tokens, and giving them equal weight. Subsequently, we conduct routing using these composite tokens enriched with prior information, as detailed below:
\begin{equation}
    \mathcal{G}^{\mathcal{P}} = \textit{top-k}(\textit{softmax}(W_g(X^{\mathcal{P}} + \ddot{\mathcal{F}}^{\mathcal{P}}))).
\end{equation}

Where $X^{\mathcal{P}}$ denote the tokens in $\mathcal{P}$, and $\ddot{\mathcal{F}}^{\mathcal{P}}$ represent their corresponding prior embeddings. Under the guidance of prior information, distribute the original tokens to the appropriate experts for processing. We obtain the final output of the PSMoE block for tokens in $\mathcal{P}$ according to Equation \ref{eq:smoe_output}, i.e., $\mathcal{G}^{\mathcal{P}}_i \cdot E_i(x^p)$. Tokens in $\mathcal{Q}$ are processed using the same method.

\subsection{PSReg}
\subsubsection{Algorithm Design.}
To solve the point cloud registration task, we propose a \textbf{P}rior-guided \textbf{S}MoE-based \textbf{Reg}istration algorithm (PSReg). Our approach builds upon \cite{yu2023peal}, and the pipeline is illustrated in Fig. \ref{fig:frame}. The framework adopts a coarse-to-fine paradigm. Initially, we utilize the KPConv-FPN backbone \cite{lin2017feature, thomas2019kpconv} network to downsample the original point clouds, obtaining superpoints and their corresponding features. Discriminative superpoint features are crucial for high-quality point cloud registration, as the accuracy of correspondences in the fine stage depends on superpoint correspondences. Therefore, our method primarily focuses on feature extraction in the coarse stage. 

We integrate our PSMoE with Transformer, leveraging PSMoE's multi-expert network to extract highly discriminative superpoint features for precise superpoint correspondence identification. Subsequently, the point matching module deduces exact point correspondences, followed by the application of local-to-global registration (LGR) \cite{qin2022geometric} for the final transformation estimation.

\subsubsection{Optimization.}
The optimization of PSReg is to minimize two correspondence supervision losses ($\mathcal{L}_c, \mathcal{L}_f$) and a load balancing loss ($\mathcal{L}_g$). The final loss function is : 
\begin{equation}
    \mathcal{L} = \mathcal{L}_c + \mathcal{L}_f + \mathcal{L}_g.
\end{equation}
The coarse correspondence loss $\mathcal{L}_c$ uses the overlap-aware circle loss \cite{qin2022geometric}, which focuses more on the positive samples with high overlap. The fine correspondence loss $\mathcal{L}_f$ uses the negative log-likelihood loss. To encourage a balanced load across experts, we add a load balancing loss $\mathcal{L}_g$, which is the same as in \cite{fedus2022switch}.
\label{sec:m3}

\section{Experiments}
We validate the efficacy of our approach on real-world datasets such as 3DMatch/3DLoMatch and synthetic datasets like ModelNet/ModelLoNet. Furthermore, we have performed comprehensive ablation experiments.

\subsection{3DMatch \& 3DLoMatch Benchmarks}

\begin{table}[t]
\centering
\resizebox{\columnwidth}{!}{
\begin{tabular}{c|ccccc|ccccc}
  \toprule
  & \multicolumn{5}{c|}{3DMatch} & \multicolumn{5}{c}{3DLoMatch} \\
  Samples & 5000 & 2500 & 1000 & 500 & 250 & 5000 & 2500 & 1000 & 500 & 250  \\
  \midrule
  \multicolumn{11}{c}{\textit{Feature Matching Recall (\%)} $\uparrow$} \\
  \midrule
  FCGF & 97.4 & 97.3 & 97.0 & 96.7 & 96.6 & 76.6 & 75.4 & 74.2 & 71.7 & 67.3 \\
  D3Feat & 95.6 & 95.4 & 94.5 & 94.1 & 93.1 & 67.3 & 66.7 & 67.0 & 66.7 & 66.5 \\
  SpinNet & 97.6 & 97.2 & 96.8 & 95.5 & 94.3 & 75.3 & 74.9 & 72.5 & 70.0 & 63.6 \\
  YOHO & 98.2 & 97.6 & 97.5 & 97.7 & 96.0 & 79.4 & 78.1 & 76.3 & 73.8 & 69.1 \\
  Predator & 96.6 & 96.6 & 96.5 & 96.3 & 96.5 & 78.6 & 77.4 & 76.3 & 75.7 & 75.3 \\
  CoFiNet & 98.1 & 98.3 & 98.1 & 98.2 & 98.3 & 83.1 & 83.5 & 83.3 & 83.1 & 82.6 \\
  GeoTr. & 97.9 & 97.9 & 97.9 & 97.9 & 97.6 & 88.3 & 88.6 & 88.8 & \textbf{88.6} & \textbf{88.3} \\
  OIF-PCR & 98.1 & 98.1 & 97.9 & 98.4 & 98.4 & 84.6 & 85.2 & 85.5 & 86.6 & 87.0 \\
  PEAL & 98.4 & 98.4 & 98.4 & 98.4 & 98.4 & 87.7 & 87.8 & 87.8 & 88.0 & 87.4 \\
  SIRA-PCR & 98.2 & 98.4 & 98.4 & 98.5 & 98.5 & \textbf{88.8} & \textbf{89.0} & \textbf{88.9} & \textbf{88.6} & 87.7 \\
  PSReg & \textbf{98.6} & \textbf{98.6} & \textbf{98.6} & \textbf{98.6} & \textbf{98.6} & 86.4 & 86.6 & 87.1 & 87.4 & 87.1 \\
  \midrule
  \multicolumn{11}{c}{\textit{Inlier Ratio (\%)} $\uparrow$} \\
  \midrule
  FCGF & 56.8 & 54.1 & 48.7 & 42.5 & 34.1 & 21.4 & 20.0 & 17.2 & 14.8 & 11.6 \\
  D3Feat & 39.0 & 38.8 & 40.4 & 41.5 & 41.8 & 13.2 & 13.1 & 14.0 & 14.6 & 15.0 \\
  SpinNet & 47.5 & 44.7 & 39.4 & 33.9 & 27.6 & 20.5 & 19.0 & 16.3 & 13.8 & 11.1 \\
  YOHO & 64.4 & 60.7 & 55.7 & 46.4 & 41.2 & 25.9 & 23.3 & 22.6 & 18.2 & 15.0 \\
  Predator & 58.0 & 58.4 & 57.1 & 54.1 & 49.3 & 26.7 & 28.1 & 28.3 & 27.5 & 25.8 \\
  CoFiNet & 49.8 & 51.2 & 51.9 & 52.2 & 52.2 & 24.4 & 25.9 & 26.7 & 26.8 & 26.9 \\
  GeoTr. & 71.9 & 75.2 & 76.0 & 82.2 & 85.1 & 43.5 & 45.3 & 46.2 & 52.9 & 57.7 \\
  OIF-PCR & 62.3 & 65.2 & 66.8 & 67.1 & 67.5 & 27.5 & 30.0 & 31.2 & 32.6 & 33.1 \\
  PEAL & 74.6 & 81.1 & 85.8 & 87.7 & 88.9 & 48.1 & 53.5 & 59.9 & 62.4 & 64.4 \\
  SIRA-PCR & 70.8 & 78.3 & 83.7 & 85.9 & 87.4 & 43.3 & 49.0 & 55.9 & 58.8 & 60.7 \\
  PSReg & \textbf{75.8} & \textbf{82.4} & \textbf{87.1} & \textbf{88.9} & \textbf{90.0} & \textbf{49.9} & \textbf{55.5} & \textbf{61.9} & \textbf{64.5} & \textbf{66.3} \\
  \midrule
  \multicolumn{11}{c}{\textit{Registration Recall (\%)} $\uparrow$} \\
  \midrule
  FCGF & 85.1 & 84.7 & 83.3 & 81.6 & 71.4 & 40.1 & 41.7 & 38.2 & 35.4 & 26.8 \\
  D3Feat & 81.6 & 84.5 & 83.4 & 82.4 & 77.9 & 37.2 & 42.7 & 46.9 & 43.8 & 39.1 \\
  SpinNet & 88.6 & 86.6 & 85.5 & 83.5 & 70.2 & 59.8 & 54.9 & 48.3 & 39.8 & 26.8 \\
  YOHO & 90.8 & 90.3 & 89.1 & 88.6 & 84.5 & 65.2 & 65.5 & 63.2 & 56.5 & 48.0 \\
  Predator & 89.0 & 89.9 & 90.6 & 88.5 & 86.6 & 59.8 & 61.2 & 62.4 & 60.8 & 58.1 \\
  CoFiNet & 89.3 & 88.9 & 88.4 & 87.4 & 87.0 & 67.5 & 66.2 & 64.2 & 63.1 & 61.0 \\
  GeoTr. & 92.0 & 91.8 & 91.8 & 91.4 & 91.2 & 75.0 & 74.8 & 74.2 & 74.1 & 73.5 \\
  OIF-PCR & 92.4 & 91.9 & 91.8 & 92.1 & 91.2 & 76.1 & 75.4 & 75.1 & 74.4 & 73.6 \\
  PEAL & 94.4 & 94.3 & 94.0 & 93.8 & 93.8 & 78.9 & 78.6 & 77.7 & 76.9 & 77.0 \\
  SIRA-PCR & 93.6 & 93.9 & 93.9 & 92.7 & 92.4 & 73.5 & 73.9 & 73.0 & 73.4 & 71.1 \\
  PSReg & \textbf{95.7} & \textbf{94.9} & \textbf{95.1} & \textbf{95.0} & \textbf{95.2} & \textbf{79.3} & \textbf{79.3} & \textbf{78.7} & \textbf{78.7} & \textbf{78.4} \\
  \bottomrule
\end{tabular}}
\caption{Quantitative results on 3DMatch and 3DLoMatch with different numbers of samples. GeoTr. denotes GeoTransformer.}
\label{tab:ransac}
\end{table}

\textbf{Dataset.} The 3DMatch \cite{zeng20173dmatch} comprises data collected from 62 indoor scenes, with 46 scenes designated for training, 8 for validation, and 8 for testing. We utilize data processed by \cite{huang2021predator}. The 3DMatch benchmark includes pairs of point clouds with an overlap rate of more than 30\%, while the 3DLoMatch benchmark only includes pairs of scans with an overlap rate of 10\%-30\%.

\noindent \textbf{Metrics.} We follow \cite{huang2021predator}, using the following three metrics to evaluate the performance of registration: Registration Recall (RR), Feature Matching Recall (FMR) and Inlier Ratio (IR).

\noindent \textbf{Registration Results.} In Tab. \ref{tab:ransac}, we compare our method against deep learning-based baselines. Specifically, we evaluated four methods based on local descriptors: FCGF \cite{choy2019fully}, D3Feat \cite{bai2020d3feat}, SpinNet \cite{ao2021spinnet}, and YOHO \cite{wang2022you}, along with six Transformer-based methods: Predator \cite{huang2021predator}, CoFiNet \cite{yu2021cofinet}, GeoTransformer \cite{qin2022geometric}, OIF-PCR \cite{yang2022one}, PEAL \cite{yu2023peal} and SIRA-PCR \cite{chen2023sira}. Following \cite{huang2021predator}, we report the performance of each method at five different numbers of correspondences (5000, 2500, 1000, 500, 250). As PEAL only reports results based on a 2D prior, while other baselines do not need 2D information, we train PEAL using a 3D prior as the input from scratch for a fair comparison. In addition, we adopted the same iterative update strategy, with 6 iterations.

RR directly reflects the final performance of the registration. As shown in Tab. \ref{tab:ransac}, our method achieved the highest RR across all sampled correspondences on 3DMatch and 3DLoMatch. Compared to the baseline GeoTransformer, our approach has improved by no less than 3.3\% on 3DMatch and no less than 5.7\% on 3DLoMatch. Under the condition of introducing the same prior information, our method has achieved a performance improvement of up to 1.5\% on 3DMatch and up to 1.8\% on 3DLoMatch compared to PEAL. Notably, our approach maintains a high RR even when the number of sampling correspondences is reduced. Especially when the number of sampling correspondences on 3DMatch is 250, we have achieved a RR of 95.2\%, which is higher than the RR of other baseline methods when the number of sampling correspondences is 5000.

As for IR, our method significantly outperforms other baseline methods, demonstrating that under the strategy of prior-guided routing, the multi-expert network of PSMoE significantly enhances the distinguishability of features, thereby obtaining reliable correspondences. For FMR, our method does not perform as well as some baseline methods on 3DLoMatch. We speculate that this is because our method is more sensitive to the quality of the priors. It can significantly improve the IR in most point cloud pairs, but in some point cloud pairs where the priors cannot provide enough correct superpoint correspondences, our method would reduce their IR, resulting in a slightly poorer FMR.

\subsection{ModelNet \& ModelLoNet Benchmarks}
\textbf{Dataset.} ModelNet40 \cite{wu20153d} consists of CAD models of 12,311 objects from 40 different categories. We follow the data settings in \cite{yew2020rpm, huang2021predator}, where the point clouds are sampled randomly from mesh faces of the CAD models, cropped and subsampled. A total of 5,112 samples are used for training, 1,202 samples for validation, and 1,266 samples for testing. Following to \cite{huang2021predator}, we evaluated two partially overlapping settings: the average overlap rate for ModelNet is 73.5\%, and for ModelLoNet, it is 53.6\%. We trained only on ModelNet and directly generalized to ModelLoNet.

\textbf{Metrics.} Following \cite{yew2020rpm, huang2021predator}, we report the Relative Rotation Error (RRE) that evaluates the error between estimated and ground truth rotation matrices, and Relative Translation Error (RTE) that measures the error between estimated and ground truth translation vectors, as well as the Chamfer distance (CD) between the registered scans.

\begin{table}[t]
\centering
\small
\begin{tabular}{c|c@{\hspace{5.2pt}}c@{\hspace{5.2pt}}c|c@{\hspace{5.2pt}}c@{\hspace{5.2pt}}c}

  \toprule
  \multirow{2}{*}{Methods} & \multicolumn{3}{c|}{ModelNet} & \multicolumn{3}{c}{ModelLoNet} \\
  & RRE & RTE & CD & RRE & RTE & CD \\
  \midrule
  DCP-v2 & 11.975 & 0.171 & 0.0117 & 16.501 & 0.300 & 0.0268 \\
  RPM-Net & \textbf{1.712} & 0.018 & \textbf{0.00085} & 7.342 & 0.124 & 0.0050 \\
  Predator & 1.739 & 0.019 & 0.00089 & 5.235 & 0.132 & 0.0083 \\
  \midrule
  \multicolumn{7}{c}{LGR} \\
  \midrule
  GeoTr. & 2.284 & 0.023 & 0.00106 & 4.176 & 0.098 & 0.0047 \\
  PEAL & 2.070 & 0.021 & 0.00100 & 4.092 & 0.108 & 0.0057 \\
  PSReg & 2.146 & 0.022 & 0.00103 & 3.952 & 0.096 & 0.0046 \\
  \midrule
  \multicolumn{7}{c}{RANSAC} \\
  \midrule
  GeoTr. & 1.875 & 0.018 & 0.00089 & 3.801 & 0.093 & 0.0047 \\
  PEAL & 1.760 & 0.018 & 0.00088 & 3.524 & 0.103 & 0.0056 \\
  PSReg & 1.745 & \textbf{0.017} & 0.00086 & \textbf{3.498} & \textbf{0.091} & \textbf{0.0043} \\
  \bottomrule
\end{tabular}
\caption{Evaluation results on ModelNet40 dataset.}
\label{tab:ModelNet40}
\end{table}

\begin{figure*}[ht]
	\centering
	\includegraphics[width=\linewidth]{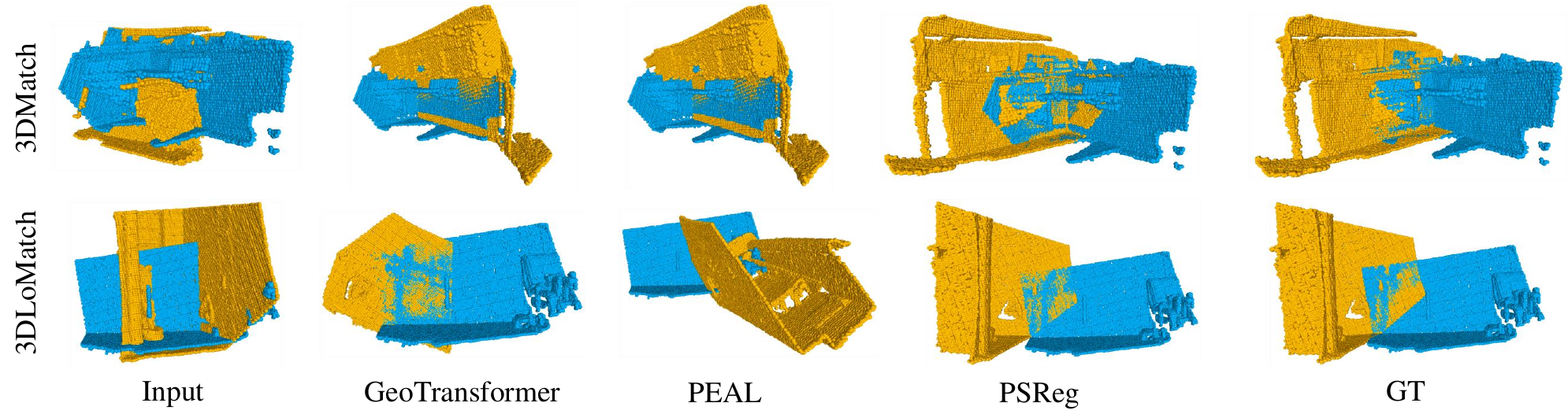}
	\caption{Visual comparison examples on 3DMatch and 3DLoMatch datasets.}
	\label{fig:sota_registration}
\end{figure*}

\textbf{Registration Results.} We compare the proposed PSReg with five SOTA baselines, including two end-to-end methods \cite{wang2019deep, yew2020rpm}, one correspondence-based method \cite{huang2021predator}, and two coarse-to-fine registration approaches \cite{qin2022geometric, yu2023peal}. For the coarse-to-fine methods, we report the metrics based on both the LGR estimator and the RANSAC estimator, with the number of RANSAC iterations set to 50K. The detailed comparison results are shown in Tab. \ref{tab:ModelNet40}. As mentioned in \cite{huang2021predator}, many end-to-end methods are specifically tuned for ModelNet40, and RPM-Net \cite{yew2020rpm} also utilizes surface normal information. Although our method's RRE and CD on ModelNet are slightly lower than RPM-Net \cite{yew2020rpm}, it surpasses other baseline methods on the more challenging ModelLoNet in terms of the RRE metric, even without using RANSAC. 

\subsection{Qualitative Results} 

Qualitative comparison with other SOTA methods is shown in Fig. \ref{fig:sota_registration}. It is worth noting that in the 3DLoMatch example, the point cloud pair only has partially overlapping walls. In such scenarios with a large number of similar structures, our method successfully achieves registration, whereas PEAL fails. From the quantitative and qualitative results, it can be seen that our proposed PSReg significantly enhances the distinguishability of features, which helps infer reliable superpoint correspondences in low-overlap point cloud registration, thereby achieving precise alignment.

\subsection{Ablation Studies}
In this section, following \cite{qin2022geometric}, we conduct extensive ablation experiments based on the LGR estimator. We report the FMR and IR of all dense point correspondences, as well as RR.

\begin{table}[h]
\centering
\small
\begin{tabular}{c|ccc|ccc}
  \toprule
  \multirow{2}{*}{Threshold ($\tau_o$)} & \multicolumn{3}{c|}{3DMatch} & \multicolumn{3}{c}{3DLoMatch} \\
  & FMR & IR & RR & FMR & IR & RR \\
  \midrule
  0.0 & \textbf{98.6} & \textbf{74.7} & \textbf{95.0} & \textbf{86.9} & \textbf{49.7} & \textbf{78.5} \\
  0.1 & 98.3 & 74.4 & 93.8 & 86.2 & 49.6 & 78.0 \\
  0.3 & 98.3 & 74.5 & 93.1 & 86.3 & 49.0 & 77.5 \\
  \bottomrule
\end{tabular}
\caption{Ablation experiments on overlap threshold.}
\label{tab:ablation0}
\end{table}

To evaluate the impact of the selection of prior superpoint correspondences, we tested various overlap thresholds, as shown in Tab. \ref{tab:ablation0}. The experimental results indicate that more prior superpoint correspondences are more conducive to registration. Hence, we ultimately set $\tau_o = 0$.

\begin{table}[h]
\centering
\small
\resizebox{\columnwidth}{!}{
\begin{tabular}{c|ccc|ccc}
  \toprule
  \multirow{2}{*}{Model} & \multicolumn{3}{c|}{3DMatch} & \multicolumn{3}{c}{3DLoMatch} \\
  & FMR & IR & RR & FMR & IR & RR \\
  \midrule
  (a) w/o SMoE & 98.4 & 73.2 & 94.2 & \textbf{88.0} & 47.1 & 77.8 \\
  (b) w/ Vanilla SMoE & 98.5 & 74.0 & 94.2 & 86.8 & 48.8 & 78.1 \\
  (c) w/ PSMoE (BC) & 98.4 & 73.9 & 94.6 & 87.8 & 49.5 & \textbf{79.2} \\
  (d) w/ PSMoE (OC) & \textbf{98.6} & \textbf{74.7} & \textbf{95.0} & 86.9 & \textbf{49.7} & 78.5 \\
  \bottomrule
\end{tabular}
}
\caption{Ablation experiments on PSReg. 'BC' means binary coding, and 'OC' means ordered coding.}
\label{tab:ablation1}
\end{table}

In Tab. \ref{tab:ablation1}, we compare different configurations of PSReg, including: (a) the baseline \cite{yu2023peal} we chose does not integrate SMoE, (b) using the vanilla SMoE \cite{fedus2022switch} on the baseline. It can be observed from Tab. \ref{tab:ablation1} that the integration of SMoE offers negligible improvement in registration performance. In addition, we also compared different encoding methods in the PCE module. According to the prior superpoint correspondences, we classify superpoints into matched and unmatched categories, encoding them with 0 and 1 respectively, which is binary coding. The ordered coding is the method shown in Fig. \ref{fig:PCE}. Tab. \ref{tab:ablation1} presents the results of different encoding methods: (c) the PCE module uses binary coding, (d) the PCE module uses ordered coding. Both encoding methods effectively improve registration performance, particularly with the ordered coding method significantly enhancing the IR.

\begin{figure}[]
	\centering
	\includegraphics[width=\linewidth]{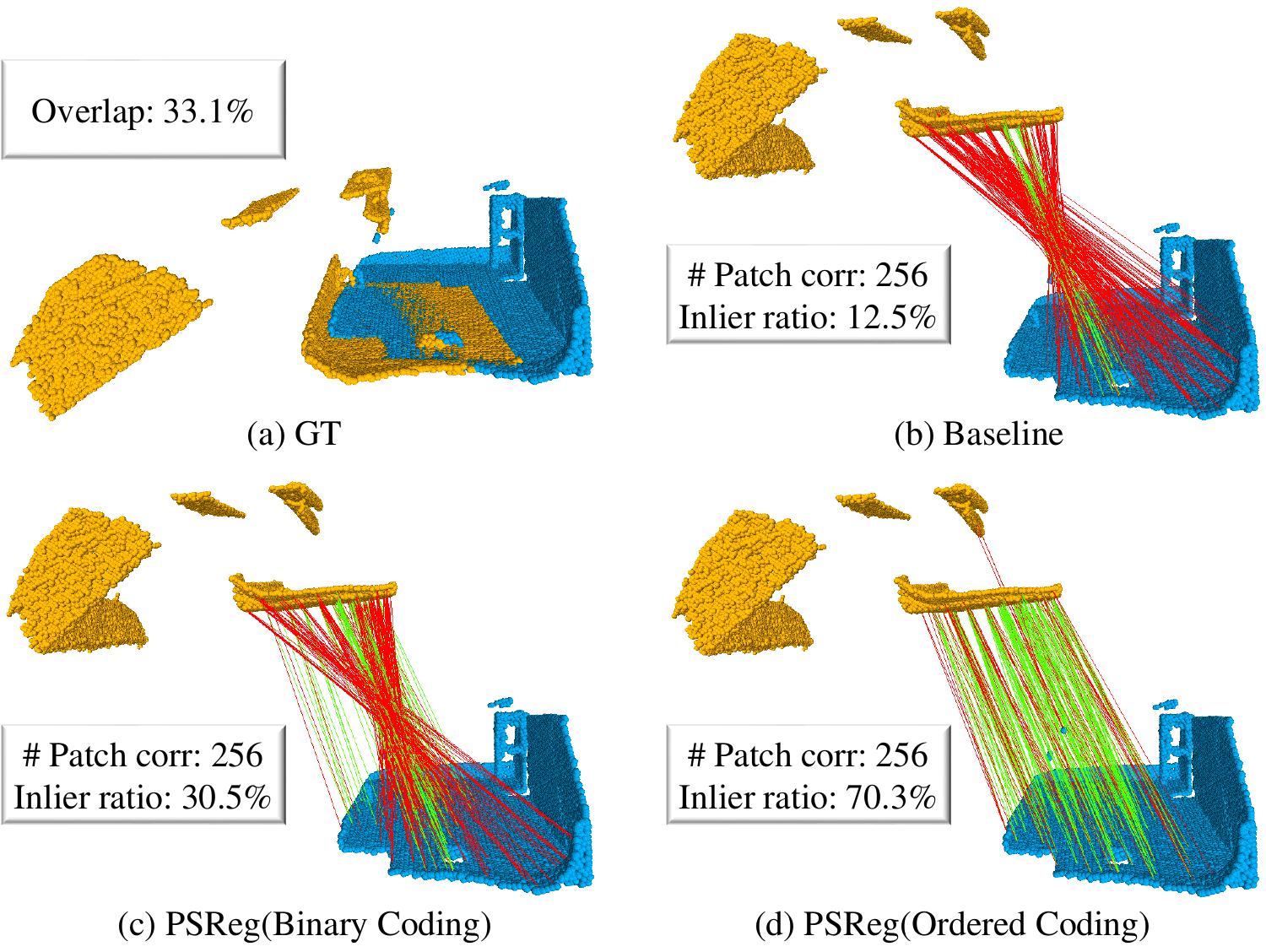}
	\caption{Visualization of superpoint correspondences with different coding methods in the PCE module.}
	\label{fig:diff_coding}
\end{figure}

Fig. \ref{fig:diff_coding} visualizes the correspondence of superpoints, it can be clearly observed that the ordered coding method achieves more accurate correspondences in overlapping areas compared to the binary coding method and the baseline. This is because the ordered coding method can distribute potentially matching tokens to the same expert as much as possible, which helps to infer more accurate correspondences.

\section{Conclusion}
In this paper, we propose the prior-guided SMoE (PSMoE) to differentiate between the ambiguous structures in the overlapping regions. Based on the PSMoE, we propose the first SMoE-based registration framework. Through extensive experiments, our method demonstrates a significant enhancement in feature discriminability, leading to a notable improvement in inlier ratio. Our method explores a new direction by utilizing a multi-expert neural network to enhance feature discriminability for point cloud registration.

\section{Acknowledgements}  
This work is supported in part by National Natural Science Foundation of China (62271237, 62132006), Jiangxi Provincial Natural Science Foundation of China (20242BAB26014) and  Science and Technology Department of Jiangxi Province (20223AEI91002).

\bibliography{references}

\end{document}